\documentclass[letterpaper, 10 pt, conference]{ieeeconf}  % Comment this line out if you need a4paper

\IEEEoverridecommandlockouts                              % This command is only needed if 
\usepackage{bm}
\usepackage{url}
\usepackage{graphicx}
\usepackage{lipsum}
\usepackage{pifont}
\usepackage{amsmath}
\usepackage{amssymb}
\usepackage{nccmath}
\usepackage{comment}
\usepackage{balance}
\usepackage{multirow}
\usepackage{textcomp}
\usepackage{multirow}
\usepackage{subcaption}
\usepackage{booktabs}
\usepackage{algorithm}
\usepackage{dblfloatfix}
\usepackage[table]{xcolor}
\usepackage[noend]{algpseudocode}

\title{\LARGE \bf
Tightly Coupled Range Inertial Localization on a 3D Prior Map \\
Based on Sliding Window Factor Graph Optimization
}

\author{Kenji Koide$^{1}$, Shuji Oishi$^{1}$, Masashi Yokozuka$^{1}$, and Atsuhiko Banno$^{1}$% <-this % stops a space
\thanks{*This work was supported in part by JSPS KAKENHI Grant Number 23K16979 and a project commissioned by the New Energy and Industrial Technology Development Organization (NEDO).}% <-this % stops a space
\thanks{$^{1}$All the authors are with the Department of Information Technology and Human Factors, the National Institute of Advanced Industrial Science and Technology, Tsukuba, Ibaraki, Japan, {\tt\small k.koide@aist.go.jp}}%
}

\begin{document}

\maketitle
\thispagestyle{empty}
\pagestyle{empty}

\setlength\floatsep{8pt}
\setlength\textfloatsep{8pt}

%%%%%%%%%%%%%%%%%%%%%%%%%%%%%%%%%%%%%%%%%%%%%%%%%%%%%%%%%%%%%%%%%%%%%%%%%%%%%%%%
\begin{abstract}

This paper presents a range inertial localization algorithm for a 3D prior map. The proposed algorithm tightly couples scan-to-scan and scan-to-map point cloud registration factors along with IMU factors on a sliding window factor graph. The tight coupling of the scan-to-scan and scan-to-map registration factors enables a smooth fusion of sensor ego-motion estimation and map-based trajectory correction that results in robust tracking of the sensor pose under severe point cloud degeneration and defective regions in a map. We also propose an initial sensor state estimation algorithm that robustly estimates the gravity direction and IMU state and helps perform global localization in 3- or 4-DoF for system initialization without prior position information. Experimental results show that the proposed method outperforms existing state-of-the-art methods in extremely severe situations where the point cloud data becomes degenerate, there are momentary sensor interruptions, or the sensor moves along the map boundary or into unmapped regions.

\end{abstract}

%%%%%%%%%%%%%%%%%%%%%%%%%%%%%%%%%%%%%%%%%%%%%%%%%%%%%%%%%%%%%%%%%%%%%%%%%%%%%%%%
\section{Introduction}

Map-based sensor localization is a crucial function for autonomous systems. Precise positional information enables the reliable navigation and recognition required for many applications, including service robots and autonomous driving. In particular, point-cloud-based localization algorithms have been one of the most popular approaches due to the recent emergence of precise and affordable range sensors, such as LiDARs and time-of-flight depth cameras.

The most naive approach to estimating a sensor pose on a 3D prior map is to iteratively apply fine point cloud registration (e.g., ICP \cite{Chetverikova} and NDT \cite{magnusson2009three}) between scan point clouds and the map point cloud. Although this naive approach works in many environments, it often becomes unreliable under aggressive sensor motion because these fine registration methods require an accurate initial guess for convergence. To improve the robustness to quick sensor motion, these scan matching methods are often integrated with additional information sources (e.g., IMU \cite{zhang2014loam} and wheel encoders \cite{Junior_2022}) to better predict the sensor pose and maintain registration results accurate. Furthermore, approaches that fuse point cloud registration errors and other sensor errors on a unified objective function (i.e., tight coupling) enable further robustness to sensor motion and partial degeneration of point clouds \cite{Xu2022}. However, it is still challenging to deal with severe point cloud degeneration and interruptions that introduce ambiguity of sensor states. Furthermore, because many existing map-based localization methods perform scan-to-map registration decoupled from ego-motion estimation, they suffer from registration failures and become corrupted at map boundaries and in regions outside the map.

\begin{figure}[tb]
  \centering
  \includegraphics[width=\linewidth]{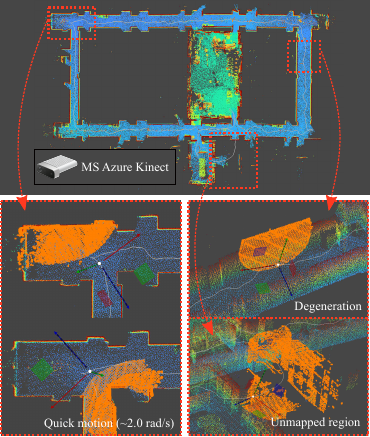}
  \caption{Indoor localization experiment using a Microsoft Azure Kinect. The proposed method enables robust and accurate pose estimation in challenging situations including under quick sensor motions, complete degeneration of point clouds, and traveling across both mapped and unmapped regions. The color of the map point cloud indicates the height of each point.}
  \label{fig:glil}
\end{figure}

In the present paper, we propose a tightly coupled range inertial localization method based on sliding window factor graph optimization by extending our previous odometry estimation algorithm \cite{Koide_2022} to a map-based localization scenario. Our formulation employs the same point cloud registration error function for scan-to-scan and scan-to-map registration constraints and jointly minimizes them along with IMU constraints in a unified objective function. This tight coupling of scan-to-scan and scan-to-map registration errors enables a smooth transition between mapped and unmapped regions. Unlike filtering-based methods, the proposed method keeps sensor states active while the states remain in a sliding window (e.g., 5 s). Furthermore, the proposed method is robust to cases where the point cloud has become degenerate and existing filtering-based methods suffer from state ambiguity. In addition, we propose a robust state initialization method that estimates the gravity direction of scan point clouds and helps perform 3- or 4-DoF global localization for initial pose estimation. Through experiments, we show that the proposed method enables robust sensor pose estimation in extremely severe situations including traveling across both mapped and unmapped regions and point clouds that have degenerated and have interruptions \footnote{See the project page for supplementary videos: \url{https://staff.aist.go.jp/k.koide/projects/icra2024_gl/}}.

This work has three main contributions:
\begin{enumerate}
    \item We propose a localization approach based on a tight coupling of scan-to-scan registration, scan-to-map registration, and IMU factors. This approach enables a smooth transition between mapped and unmapped regions and makes pose estimation robust to point cloud degeneration and interruptions.
    \item We develop a simple and robust gravity direction estimation method based on the batch optimization of sensor poses and IMU measurements. This method allows us to perform global localization in a reduced DoF.
    \item We release an evaluation dataset that can be used to evaluate the robustness of map-based localization algorithms in extremely severe situations.
\end{enumerate}

%%%%%%%%%%%%%%%%%%%%%%%%%%%%%%%%%%%%%%%%%%%%%%%%%%%%%%%%%%%%%%%%%%%%%%%%%%%%%%%%
\section{Related Work}

%%%%%%%%%%%%%%%%%%%%%%%%%%%%%%%%%%%%%
\subsection{Iterative Scan Matching}

Estimation of the position of a sensor on a map is essential for navigation systems. While several kinds of map representations are used, depending on the use scenario (e.g., high-definition vector maps \cite{Ma_2019} and wiki-based open geographic maps like OpenStreetMap \cite{Yan_2019, Cho_2022}), 3D point cloud maps are among the most popular representations owing to their simplicity and expressiveness. Because constructing a point cloud map is relatively easy with recent precise range sensors and mapping algorithms, point cloud maps are used for a wide range of applications, from indoor service robots to outdoor driving of autonomous vehicles.

For sensor localization on a point cloud map, local point cloud registration methods, such as ICP \cite{Chetverikova} and NDT \cite{magnusson2009three}, are often used. By iteratively applying point cloud registration between scan points and map points, we can easily and efficiently track a 6-DoF sensor pose. A key to obtaining fine registration results is to provide an accurate initial guess to ensure the convergence. For this reason, additional information sources, for example, IMU \cite{zhang2014loam}, wheel encoders \cite{Junior_2022}, and leg joint angles of quadruped robots \cite{Wisth2023}, are jointly used with range sensors to compensate for sensor motion and better predict the current sensor pose to be used as an initial guess for point cloud registration. A recent trend in sensor motion estimation is a tight coupling of multi-sensor constraints that fuses the cost functions of several sensors on a unified objective function. This makes it possible to keep the system well constrained when the data of a sensor become unreliable. In particular, the LiDAR-IMU tight coupling approach has been actively explored in recent studies \cite{Bai2022,Shan_2021,Nguyen_2021}. Because the tight coupling approach significantly increases the computation cost, lightweight iterated Kalman filters are often used \cite{Xu2022}. However, these filtering-based approaches suffer from severe degeneration of point cloud data because they immediately marginalize past frames when a new frame arrives and cannot accurately propagate the uncertainty of past observations. Furthermore, in many studies, scan-to-map registration was decoupled from the odometry estimation algorithm, which results in difficulty correcting sensor poses at map boundaries and maintaining system stability outside the map.

%%%%%%%%%%%%%%%%%%%%%%%%%%%%%%%%%%%%%
\subsection{Monte Carlo Localization}

Monte Carlo localization (MCL), which represents and estimates a state distribution with a finite set of state samples, is a popular approach to 2D map-based localization \cite{Fox_2003, probrobo}. Owing to its expressiveness for non-linear and non-Gaussian distributions, it is extremely robust to observation ambiguity and multi-hypothesis situations. Despite its success in 2D localization, MCL has not been commonly used in 3D localization problems because the required number of samples to fill a unit space grows exponentially as the number of dimensions increases, which becomes a computational burden. Many studies have performed 3D MCL in 3-DoF \cite{Saarinen_2013} or 4-DoF \cite{Perez_Grau_2017} with assumptions on sensor motion and environment structure. Although several studies have tackled 6-DoF MCL with a smaller number of samples using efficient state sampling \cite{Akai_2020,Deng_2021,Maken_2022}, it is still difficult to perform a general full 6-DoF MCL without such assumptions and a prior knowledge of the sensor and the environment.

%%%%%%%%%%%%%%%%%%%%%%%%%%%%%%%%%%%%%%%%%%%%%%%%%%%%%%%%%%%%%%%%%%%%%%%%%%%%%%%%
\section{Methodology}

\begin{figure*}[tb]
  \centering
  \includegraphics[width=0.8\linewidth]{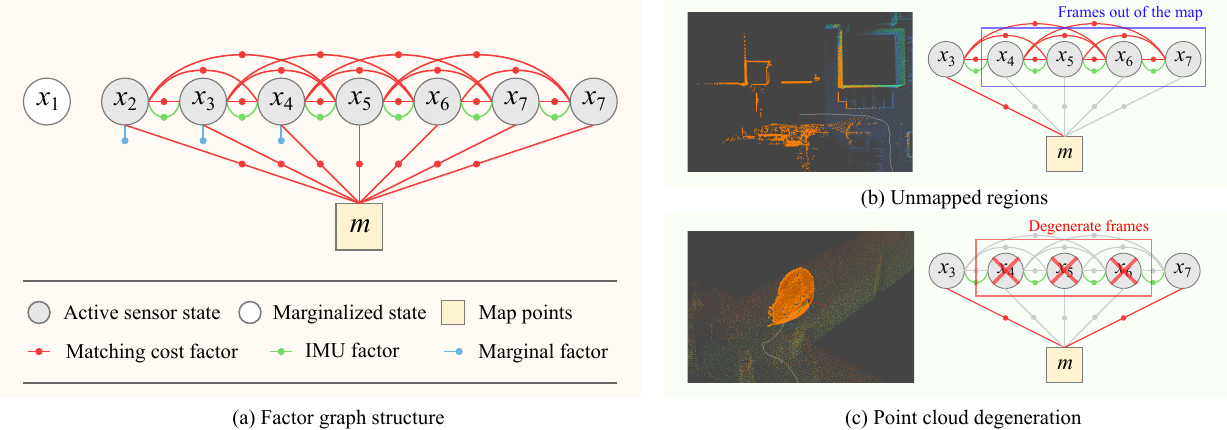}
  \caption{Proposed factor graph structure. (a) The proposed factor graph consists of tightly coupled scan-to-scan and scan-to-map registration factors along with IMU factors. Old frames that leave the optimization window are marginalized to bound the computation cost. The proposed method enables robust pose estimation in challenging situations including (b) the sensor travels over unmapped places, and (c) point cloud data becomes degenerated. }
  \label{fig:graph_extreme}
\end{figure*}

%%%%%%%%%%%%%%%%%%%%%%%%%%%%%%%%%%%%%
\subsection{Notation}

Given a map point cloud $\mathcal{M} = \{ {\bm p}_k^M \in \mathbb{R}^3 \mid_{k = 1, \ldots, N^M} \}$, we estimate a time series of sensor states ${\bm x}_t$ in the map frame using measurements of point cloud scans $\mathcal{P}_t = \{ {\bm p}_k^S \in \mathbb{R}^3 \mid_{k = 1, \ldots, N^P} \}$ and IMU data $\mathcal{I}_t = [{\bm a}_t, {\bm \omega}_t]$, where ${\bm a}_t \in \mathbb{R}^3$ and ${\bm \omega}_t \in \mathbb{R}^3$ are the linear acceleration and the angular velocity, respectively. The sensor state to be estimated is defined as:
\begin{align}
{\bm x}_t = [ {\bm T}_t, {\bm v}_t, {\bm b}_t ],
\end{align}
where ${\bm T}_t = [{\bm R}_t \mid {\bm t}_t] \in SE(3)$ and ${\bm v} \in \mathbb{R}^3$ are respectively the sensor pose and velocity in the map frame, and ${\bm b}_t = [{\bm b}_t^a, {\bm b}_t^\omega] \in \mathbb{R}^6$ is the IMU acceleration and angular velocity bias. Note that we transform scan points into the IMU coordinate frame and consider them as if they are in the same coordinate frame for efficiency and simplicity.

In the following Sec. \ref{sec:matching_factor} and \ref{sec:imu_factor}, we describe the building blocks of the proposed framework, namely the point cloud registration factor and the preintegrated IMU factor, and then present the proposed factor graph structure in Sec. \ref{sec:factor_graph}.

%%%%%%%%%%%%%%%%%%%%%%%%%%%%%%%%%%%%%
\subsection{Point Cloud Registration Factor}
\label{sec:matching_factor}

To constrain the relative pose between point clouds $\mathcal{P}_i$ and $\mathcal{P}_j$, we use the voxelized GICP (VGICP) registration error factor with GPU acceleration \cite{Koide_2022}. As with GICP \cite{segal2009generalized}, VGICP models each point as a Gaussian distribution ${\bm p}_k \sim \mathcal{N}({\bm \mu}_k, {\bm C}_k)$ representing the local geometrical shape and computes a sum of the distribution-to-distribution distances between corresponding points as follows:
\begin{align}
\label{eq:epc}
e^\text{PC} ( \mathcal{P}_i, \mathcal{P}_j, {\bm T}_i, {\bm T}_j ) &= \sum_{{\bm p}_k \in \mathcal{P}_i} e^\text{D2D} ({\bm p}_k, {\bm T}_i^{-1} {\bm T}_j), \\
\label{eq:d2d}
e^\text{D2D}({\bm p}_k, {\bm T}_{ij}) &= {\bm d}_k^\top \left( {\bm C}'_k + {\bm T}_{ij} {\bm C}_k {\bm T}_{ij}^\top \right)^{-1} {\bm d}_k,
\end{align}
where ${\bm T}_{ij} = {\bm T}_i^{-1} {\bm T}_j$ is the relative transformation between $\mathcal{P}_i$ and $\mathcal{P}_j$, ${\bm p}'_k \sim \mathcal{N}( {\bm \mu}'_k, {\bm C}'_k)$ is the point in $\mathcal{P}_j$ nearest to ${\bm p}_k$, and ${\bm d}_k = {\bm \mu}'_k - {\bm T}_{ij} {\bm \mu}_k$ is the residual between ${\bm \mu}'_k$ and ${\bm \mu}_k$. The covariance matrix ${\bm C}_k$ is computed from neighboring points of ${\bm p}_k$. To avoid a costly nearest neighbor search, VGICP voxelizes the target point cloud $\mathcal{P}_j$ at a specific resolution $r$ and stores the average of means and covariance matrices of points in each voxel. During cost evaluation, it looks up a voxel corresponding to each input point and computes Eq. \ref{eq:d2d} using the voxel as ${\bm p}'_k$.

%%%%%%%%%%%%%%%%%%%%%%%%%%%%%%%%%%%%%
\subsection{Preintegrated IMU Factor}
\label{sec:imu_factor}

To efficiently incorporate IMU measurements into the factor graph, we use the preintegration technique \cite{Forster_2015}. Given an IMU measurement (${\bm a}_t$ and ${\bm \omega}_t$), the sensor states evolves over time as follows:
\begin{align}
\label{eq:imu_evol_R}
{\bm R}_{t + \Delta t} &= {\bm R}_t \exp \left( \left( {\bm \omega}_t - {\bm b}_t^{\omega} - {\bm \eta}_k^{\omega} \right) \Delta t \right), \\
\label{eq:imu_evol_v}
{\bm v}_{t + \Delta t} &= {\bm v}_t + {\bm g} \Delta t + {\bm R}_t \left( {\bm a}_t - {\bm b}_t^a - {\bm \eta}_t^a \right) \Delta t, \\
\label{eq:imu_evol_p}
{\bm t}_{t + \Delta t} &= {\bm t}_t + {\bm v}_t \Delta t + \frac{1}{2} {\bm g} \Delta t^2 + \frac{1}{2} {\bm R}_t \left( {\bm a}_t - {\bm b}_t^a - {\bm \eta}_t^a \right) \Delta t^2,
\end{align}
where ${\bm g}$ is the gravity vector and ${\bm \eta}_t^a$ and ${\bm \eta}_t^{\omega}$ represent white noise in the IMU measurement.

The IMU preintegration factor integrates Eqs. \ref{eq:imu_evol_R} -- \ref{eq:imu_evol_p} between times $i$ and $j$ to obtain the relative sensor motion $\Delta {\bm R}_{ij}$, $\Delta {\bm t}_{ij}$, and $\Delta {\bm v}_{ij}$ \cite{Forster_2015}. Then, the IMU prediction error is calculated as follows:
\begin{align}
\begin{split}
\label{eq:preint_R}
e^{\text{\it IMU}} ( {\bm x}_i, {\bm x}_j ) &= \| \log \left( \Delta {\bm R}_{ij}^\top {\bm R}_i^\top {\bm R}_j \right) \|^2 \\
+ & \| \Delta {\bm t}_{ij} - {\bm R}_i^\top \left( {\bm t}_j - {\bm t}_i - {\bm v} \Delta t_{ij} - \frac{1}{2} {\bm g} \Delta t_{ij}^2 \right) \|^2 \\
+ & \| \Delta {\bm v}_{ij} - {\bm R}_i^\top \left( {\bm v}_j - {\bm v}_i -{\bm g} \Delta t_{ij} \right) \|^2 .
\end{split}
\end{align}
Because IMU measurements provide a constant amount of sensor motion information independent of the environment structure, they help keep the factor graph well-constrained in environments where point cloud registration factors can be degenerate (e.g., tunnels and long corridors).

%%%%%%%%%%%%%%%%%%%%%%%%%%%%%%%%%%%%%
\subsection{Factor Graph Structure}
\label{sec:factor_graph}

Fig. \ref{fig:graph_extreme} (a) illustrates the proposed factor graph structure, which consists of four major elements: scan-to-scan registration factors, scan-to-map registration factors, IMU factors, and marginal factors.

{\bf Ego-motion estimation:} To estimate the sensor ego-motion, scan-to-scan registration factors are created between scan point clouds. To make the estimation robust to quick sensor motion, we create scan-to-scan registration factors between the latest frame and $N^\text{pre}$ preceding frames (e.g., 3 frames), which results in a densely connected factor graph. IMU factors are created between consecutive frames to help predict the sensor pose and keep the factor graph well-constrained under point cloud degeneration.

{\bf Map-based drift correction:} To suppress estimation drift, we create scan-to-map registration factors between every frame and the map point cloud. We treat the map point cloud as a static frame and create unary registration factors that constrain only the frame poses. The key idea here is to use the same error function for scan-to-scan and scan-to-map registration factors and jointly minimize them. This structure makes it possible to maintain sensor pose tracking on map boundaries and even in unmapped regions. As shown in Fig. \ref{fig:graph_extreme} (b), even if a frame has only partial overlap with the map and the scan-to-map registration factor becomes degenerate, the latest frame is still well-constrained by the scan-to-scan registration factors, and we can safely incorporate the degenerate scan-to-map registration errors into the factor graph. Furthermore, when there is absolutely no overlap with the map, the factor graph simply falls back to a range inertial odometry estimation and maintains sensor pose tracking. Once the sensor comes back to the map, scan-to-map registration errors are re-activated and smoothly correct the estimation drift while retaining the consecutive frame matching consistency. The proposed graph structure also makes the estimation robust to degeneration of point cloud data. As in Fig. \ref{fig:graph_extreme} (c), the tightly coupled scan-to-map registration factors enable deficient relative pose constraints (5-DoF with ambiguity in the corridor direction in the shown case) to be incorporated into the factor graph to reduce estimation errors.

{\bf Optimization:} To bound the computation cost, we marginalize old frames that leave the optimization window (e.g., 5 s) and add constant linear factors to compensate for the marginalized factors at the last evaluation point. For factor graph optimization and marginalization, we use the iSAM2 optimizer \cite{Kaess_2011} and its efficient Bayes tree elimination implemented in GTSAM \cite{gtsam}.

In summary, the objective function of the proposed framework is defined as follows:
\begin{align}
\begin{split}
e(\mathcal{X}) 
& = \sum_{{\bm x}_i \in \mathcal{X}} \sum_{j=i - N^\text{pre}}^{i - 1} e^\text{PC} ( \mathcal{P}_i, \mathcal{P}_j, {\bm T}_i, {\bm T}_j ) \\
& + \sum_{{\bm x}_i \in \mathcal{X}} e^\text{PC} ( \mathcal{P}_i, \mathcal{M}, {\bm T}_i, {\bm I}^{6 \times 6} ) \\
& + \sum_{{\bm x}_i \in \mathcal{X}} e^\text{IMU} ( {\bm x}_{i - 1}, {\bm x}_i ) + \mathcal{C},
\end{split}
\end{align}
where $\mathcal{X}$ is a set of sensor states in the optimization window and $\mathcal{C}$ is a set of linear factors for marginalization.

%%%%%%%%%%%%%%%%%%%%%%%%%%%%%%%%%%%%%
\subsection{Gravity Direction Estimation}

In practice, the gravity direction is useful information that allows aligning the upward directions of scan and map point clouds and enables the search space to be narrowed down for global localization in 4-DoF \cite{Lim_2022} or 3-DoF with an additional sensor height assumption \cite{Kim_2018}. Although the gravity direction can easily be obtained using linear acceleration data when the sensor is stationary, the estimated gravity direction can be affected by sensor motion.

To aid global localization used for initial pose estimation, we developed a simple and robust gravity direction and IMU state estimation method based on batch optimization. We first estimate the sensor trajectory using only point cloud data. To this end, we use a combination of the continuous time ICP (CT-ICP) \cite{Dellenbach2022} and the voxel-based points container structure (linear iVox) \cite{Bai2022}. Given the estimated sensor trajectory and IMU measurements in a certain optimization window (e.g., 2 s), we create a factor graph that consists of relative pose factors based on the estimated trajectory and preintegrated IMU factors between consecutive frames. The objective function is defined as follows:
\begin{align}
\label{eq:init}
e^\text{Init} (\mathcal{X}) &= \sum_{ {\bm x}_i \in \mathcal{X} } \left( e^\text{RP} ( {\bm x}_{i - 1}, {\bm x}_i ) + e^\text{IMU} ({\bm x}_{i - 1}, {\bm x}_i) \right),
\end{align}
where $e^\text{RP} ({\bm x}_i, {\bm x}_j) = \| \log ( \widetilde{\bm T}_{ij}^{-1} {\bm T}_i^{-1} {\bm T}_j ) \|^2$ is the relative pose constraint based on the relative pose measurement $\widetilde{\bm T}_{ij}$ computed from the estimated sensor trajectory. Because this objective function has ambiguity in the translation and yaw-axis rotation, we add a large constant to the diagonal elements of the information matrix of the first pose to fix the gauge freedom and maintain system positive definiteness. As an initial estimate of the very first frame, we compute the average linear acceleration vector and rotate the first frame such that the average linear acceleration vector is aligned with the world gravity direction. The initial rotation of the first frame is given by
\begin{align}
{\bm \psi} &= {\bm \mu}^a \times {\bm g}^w, \\
{\bm R}_0 &= \mathbb{I}^{6 \times 6} + \widehat{{\bm \psi}} + \widehat{\bm \psi}^2 \frac{ (1 - {\bm \mu}^a \cdot {\bm g}^w) }{ | {\bm \psi} | ^ 2 },
\end{align}
where ${\bm \mu}^a$ is the normalized average linear acceleration vector, ${\bm g}^w = [0, 0, 1]^T$ is the gravity direction in the world frame, and $\widehat{\cdot}$ is the hat operator to compute the skew symmetric matrix. We use the Levenberg-Marquardt optimizer to minimize Eq. \ref{eq:init}. After optimization, we use the optimized sensor pose, velocity, and IMU bias of the last frame as the initial state of the localization system.

To demonstrate the usefulness of the gravity direction information for automatic system initialization, we implemented a 2D occupancy gridmap-based global localization algorithm using branch-and-bound search \cite{Hess_2016} \footnote{The implementation of the 2D global localization is available at: \url{https://github.com/koide3/hdl_global_localization}}. In outdoor experiments shown in Sec. \ref{sec:outdoor}, we obtain a 2D slice of scan point clouds using the estimated gravity direction and feed it to the global localization algorithm to obtain an initial position estimate. We then initialize the proposed localization framework with the estimated initial position. Note that we designed the framework so that it can dynamically load a global localization module from a shared library. The framework is agnostic to the global localization algorithm and any 3- or 4-DoF global localization algorithm (e.g., \cite{Lim_2022}) can easily be incorporated into the proposed framework.

%%%%%%%%%%%%%%%%%%%%%%%%%%%%%%%%%%%%%%%%%%%%%%%%%%%%%%%%%%%%%%%%%%%%%%%%%%%%%%%%
\section{Experiment}

%%%%%%%%%%%%%%%%%%%%%%%%%%%%%%%%%%%%%
\subsection{Indoor Experiment}
\label{sec:indoor}

% Easy01
% glil          : 0.054304 +- 0.007990
% fast_lio_odom : 2.484667 +- 1.216373
% fast_lio_loc  : 0.068420 +- 0.043713
% hdl_loc       : 0.210297 +- 0.177167

% Easy02
% glil          : 0.041350 +- 0.011302
% fast_lio_odom : 7.101489 +- 2.395595
% fast_lio_loc  : 0.149634 +- 0.117944
% hdl_loc       : 15.048311 +- 10.505571

% Hard01
% glil          : 0.282799 +- 0.253327
% fast_lio_odom : 27.941562 +- 23.022280
% fast_lio_loc  : 27.265159 +- 25.352512
% hdl_loc       : 25.507035 +- 24.431026

\newcommand{\cmark}{\ding{51}}%
\newcommand{\xmark}{\ding{55}}%
\setlength{\tabcolsep}{5pt}

{\bf Experimental setting:} To demonstrate the robustness of the proposed method, we conducted localization experiments in the indoor environment shown in Fig. \ref{fig:glil}. Using a Microsoft Azure Kinect, we recorded two sequences (Easy01 and Easy02) of point cloud and IMU data without aggressive motion, and another one sequence (Hard) with quick sensor motion, point cloud degeneration, and traveling across both mapped and unmapped regions. The durations of the sequences Easy01, Easy02, and Hard were respectively 139, 136, and 164 s \footnote{The dataset is available at \url{https://staff.aist.go.jp/k.koide/projects/icra2024_gl/}}.

As a baseline, we ran two localization algorithms: FAST\_LIO\_LOCALIZATION (FAST\_LIO\_LOC) \footnote{\url{https://github.com/HViktorTsoi/FAST_LIO_LOCALIZATION}} and hdl\_localization \cite{Koide_2019}. FAST\_LIO\_LOC uses FAST\_LIO2, a tightly coupled LiDAR-IMU odometry estimation based on an iterated Kalman filter \cite{Xu2022}, to estimate the sensor ego-motion and periodically performs scan-to-map registration to correct estimation drift. For comparison, we also ran FAST\_LIO2 \cite{Xu2022} without map-based pose correction. hdl\_localization \cite{Koide_2019} performs NDT-based scan-to-map registration and IMU fusion based on an unscented Kalman filter. For all the evaluated methods, we provided an initial pose manually.

To obtain reference sensor trajectories, we manually aligned scan point clouds with the map point cloud and performed batch optimization of the GICP scan-to-map registration errors and IMU motion errors. We evaluated the estimated trajectories with the absolute trajectory error (ATE) metric \cite{Zhang_2018} using the {\it evo} toolkit \footnote{\url{https://github.com/MichaelGrupp/evo}}.

\begin{table}[tb]
  \caption{Absolute trajectory errors for indoor sequences}
  \label{tab:results_easy}
  \centering
  \scriptsize
  \begin{tabular}{l|ccc}
  \toprule
                    & \multicolumn{3}{c}{ATE [m]} \\
  Method            & Easy01                & Easy02                  & Hard                  \\
  \midrule
  FAST\_LIO (odom)  & 2.485 $\pm$ 1.216     & 7.101 $\pm$ 2.396       & 27.942 $\pm$ 23.022   \\
  FAST\_LIO\_LOC    & 0.068 $\pm$ 0.044     & 0.150 $\pm$ 0.118       & 27.265 $\pm$ 25.353   \\
  hdl\_localization & 0.210 $\pm$ 0.177     & 15.048 $\pm$ 10.506     & 25.507 $\pm$ 24.431   \\
  Proposed          & \bf 0.054 $\pm$ 0.008 & \bf 0.041 $\pm$ 0.011   & \bf 0.282 $\pm$ 0.253 \\
  \bottomrule
  \end{tabular}
\end{table}

{\bf Easy sequences:} Table \ref{tab:results_easy} summarizes the ATEs of the evaluated methods. We can see that FAST\_LIO without map-based correction showed large trajectory estimation errors for the Easy01 and Easy02 sequences (2.485 and 7.101 m) because the estimation drift quickly accumulated due to the small feature-less environment and the narrow field of view of the sensor. FAST\_LIO\_LOC significantly improved the ATEs (0.068 and 0.150 m), and this result suggests the necessity of map-based correction to maintain the accuracy of sensor pose tracking. Although hdl\_localization showed a decent ATE for the Easy01 sequence (0.210 m), its accuracy largely deteriorated for the Easy02 sequence (15.048 m) because of a scan matching failure caused by the small feature-less environment. The proposed method showed the best ATEs for the Easy01 and Easy02 sequences (0.054 and 0.041 m) thanks to its robustness to a feature-less environment.

\begin{figure}[tb]
  \centering
  \includegraphics[width=0.8\linewidth]{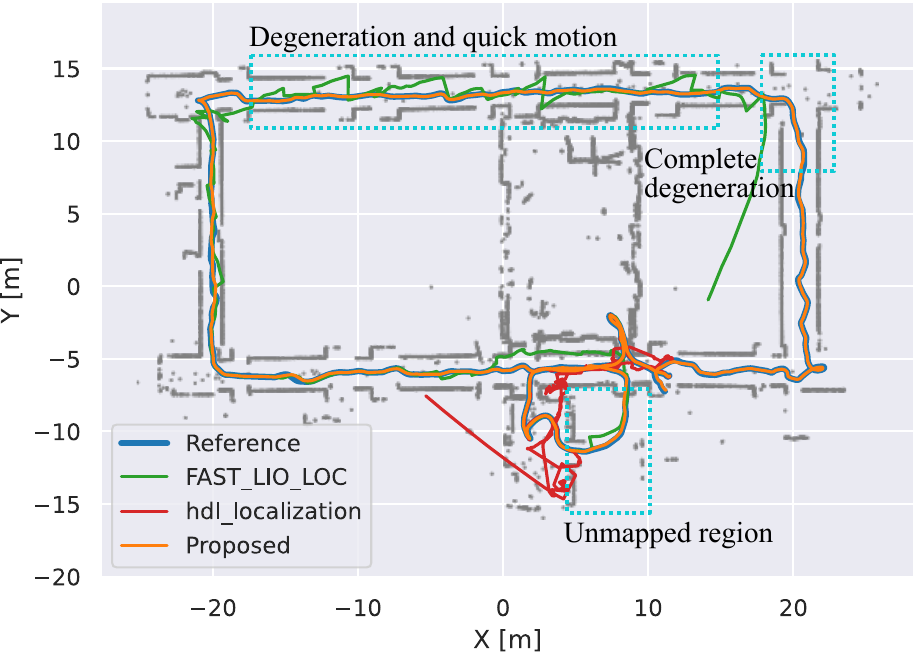}
  \caption{Estimated trajectories for the Hard sequence. Existing methods suffered from unmapped regions and severe degeneration of point clouds. The proposed method successfully continued tracking the sensor pose under these severe conditions thanks to its tightly coupled registration factors and windowed state optimization.}
  \label{fig:traj_hard}
\end{figure}

{\bf Hard sequence:} For the Hard sequence, both FAST\_LIO\_LOC and hdl\_localization became corrupted. Fig. \ref{fig:traj_hard} shows the estimated trajectories. Because hdl\_localization largely relied on scan-to-map registration to maintain sensor pose tracking, it immediately became corrupted when the sensor entered an unmapped region. FAST\_LIO\_LOC showed a slight estimation error in the unmapped region because of the decoupled scan-to-map registration that became unreliable when the overlap with the map was small. We observed that FAST\_LIO\_LOC became unstable under point cloud degeneration because the underlying FAST\_LIO2 suffered from pose ambiguity that made it difficult to accurately accumulate scan points into the model point cloud. As a consequence, it eventually became corrupted when a complete degeneration of point clouds occurred. The proposed method successfully continued tracking the sensor pose during the Hard sequence and showed the best ATE among the evaluated methods (0.282 m). Because of the tightly coupled scan-to-scan and scan-to-map registration factors, it showed smooth estimation over the unmapped region. Furthermore, its windowed optimization made it possible to deal with complete degeneration of point cloud data that resulted in a smooth trajectory estimation result.

%%%%%%%%%%%%%%%%%%%%%%%%%%%%%%%%%%%%%
\subsection{Outdoor Experiment}
\label{sec:outdoor}

\begin{figure}[tb]
  \centering
  \includegraphics[width=1.0\linewidth]{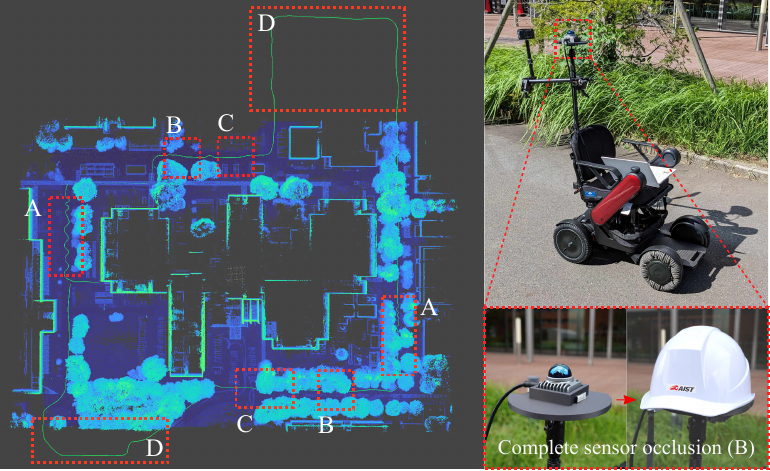}
  \caption{Outdoor experimental environment (280 $\times$ 200 $\text{m}^2$). The evaluation sequences include four challenging situations: A) large translational and rotational movements, B) point cloud data interruptions, C) aggressive sensor rotations, and D) traveling over unmapped regions. In particular, in the B regions, the sensor was completely occluded several times, which created extreme challenges to localization methods.}
  \label{fig:outdoor_environment}
\end{figure}

\renewcommand{\labelenumi}{\Alph{enumi})}

{\bf Experimental setting:} We conducted experiments in the outdoor environment shown in Fig. \ref{fig:outdoor_environment}. We recorded two sequences of point cloud and IMU data using a Livox MID360. To evaluate the robustness of localization methods, we included the following challenging situations in each sequence:
\begin{enumerate}
  \item The sensor was moved with quick translation and rotation ($\sim$1.6 m/s and $\sim$1.1 rad/s).
  \item The sensor view was occluded completely so that point cloud data were interrupted several times.
  \item The sensor was strongly shaken in random directions (5.0 rad/s) over a long interval (5--10 s).
  \item The sensor moved outside the map and traveled in unmapped regions for about 150--300 m.
\end{enumerate}
Because the dataset constrains extremely difficult situations, when a localization method becomes corrupted, we restart that method and count the number of corruptions that occurred in each sequence. 

For the proposed method, we did not provide an initial sensor pose but rather used a combination of the proposed gravity direction estimation and the 2D global localization \cite{Hess_2016} for automatic system initialization.

\begin{figure}[tb]
  \centering
  \includegraphics[width=1.0\linewidth]{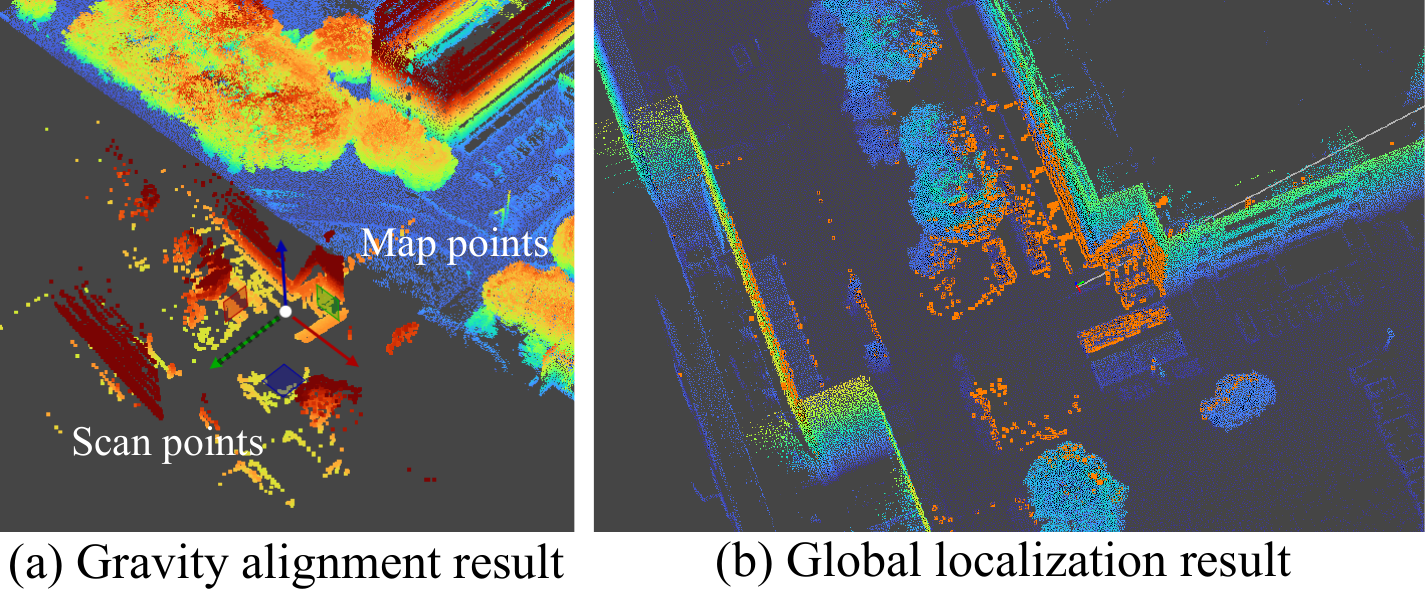}
  \caption{Automatic initialization result. (a) The gravity direction of the LiDAR scans was estimated, and (b) the 2D global localization successfully estimated the initial sensor position.}
  \label{fig:initialization}
\end{figure}

{\bf Initialization: } For both the Outdoor01 and Outdoor02 sequences, the initialization process properly estimated the gravity direction of scan points as shown in Fig. \ref{fig:initialization} (a) although the LiDAR was slightly tilted. Then, the 2D global localization was performed based on the gravity-aligned scan points, and it successfully estimated the sensor position, and the localization process was initiated. The gravity direction estimation and global localization took approximately 2.0 and 3.5 s, respectively. Consequently, the automatic initialization process took 5.5 s in total.

% initialization : 2 sec
% gloc start : 22:01:32.659   end : 22:01:36.190

\begin{figure}[tb]
  \centering
  \includegraphics[width=0.75\linewidth]{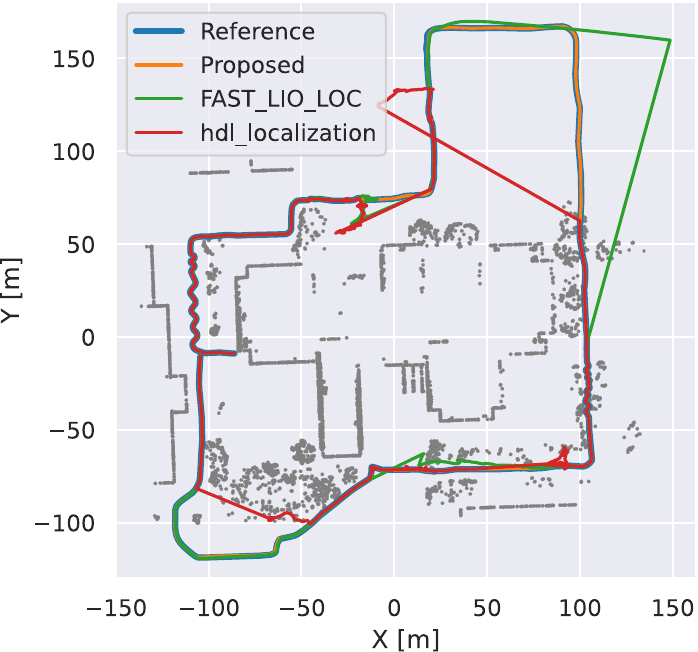}
  \caption{Estimated trajectories for the Outdoor01 sequence.}
  \label{fig:traj_outdoor}
\end{figure}

% outdoor01
% glil         : 0.958675 +- 0.838494  0 corruptions
% fast_lio_loc : 6.983963 +- 6.692931  4 corruptions
% hdl_loc      : 5.592192 +- 5.252291  5 corruptions
% 
% outdoor02
% glil         : 0.596935 +- 0.401234  0 corruptions
% fast_lio_loc : 4.848815 +- 4.692089  4 corruptions
% hdl_loc      : 7.136962 +- 6.832634  5 corruptions

% preprocess  : 7.732026 +- 1.889104 [msec]
% optimization: 9.483660 +- 6.101649 [msec]

\begin{table}[tb]
  \caption{Absolute trajectory errors for outdoor sequences}
  \label{tab:results_outdoor}
  \centering
  \scriptsize
  \begin{tabular}{l|cc|cc}
  \toprule
  \multirow{2}{*}{Method}
                    & \multicolumn{2}{c|}{Outdoor01} &  \multicolumn{2}{c}{Outdoor02} \\
                    & ATE [m]          & Corruptions & ATE [m]           & Corruptions \\
  \midrule
  FAST\_LIO\_LOC    & 6.983 $\pm$ 6.693     & 4      & 4.849 $\pm$ 4.692     & 4       \\
  hdl\_localization & 5.592 $\pm$ 5.252     & 5      & 7.137 $\pm$ 6.833     & 5       \\
  Proposed          & \bf 0.959 $\pm$ 0.838 & \bf 0  & \bf 0.597 $\pm$ 0.401 & \bf 0   \\
  \bottomrule
  \end{tabular}
\end{table}

{\bf Estimation result: } Table \ref{tab:results_outdoor} summarizes the quantitative evaluation results, and Fig. \ref{fig:traj_outdoor} shows the estimated trajectories. Owing to IMU fusion, all the methods were able to continue tracking the sensor pose under quick translation and rotation in the region (A). Under data interruptions, both the existing methods became unstable, and hdl\_localization immediately became corrupted in the region (B). Although FAST\_LIO\_LOC maintained pose tracking in the region (B), in the following region (C), it became corrupted due to very aggressive sensor rotation. Furthermore, both methods failed to maintain sensor pose tracking in unmapped regions (D). This result shows the weakness of decoupled scan-to-map registration, which can be unreliable in places where only a small overlap with the map is available.

The proposed method successfully continued tracking the sensor pose through the Outdoor01 and Outdoor02 sequences without estimation corruptions, whereas FAST\_LIO\_LOC and hdl\_localization respectively became corrupted 4 and 5 times for each sequence. This result suggests the robustness of the proposed method owing to the sliding-window-based optimization and tightly coupled scan-to-map registration factors.

{\bf Processing time: } For the proposed method, point cloud preprocessing and factor graph optimization respectively took 7.73 $\pm$ 1.89 and 9.48 $\pm$ 6.10 ms, and the total processing time per frame was 17.21 ms (58.1 FPS), which was much faster than the real-time requirement (10 FPS). 

%%%%%%%%%%%%%%%%%%%%%%%%%%%%%%%%%%%%%%%%%%%%%%%%%%%%%%%%%%%%%%%%%%%%%%%%%%%%%%%%
\section{Conclusion}

In this work, we proposed a map-based localization algorithm using the tight coupling of scan-to-scan and scan-to-map registration factors and IMU factors. Sensor states in a sliding window were continuously optimized. The proposed approach enabled dealing with severe point cloud degeneration and traveling across both mapped and unmapped regions. Through indoor and outdoor experiments, the proposed method was shown to successfully continue tracking the sensor pose in challenging situations with a bounded computation cost.

In future work, we plan to incorporate a more sophisticated 4-DoF global localization method \cite{Aoki2024} for reliable system initialization in complex 3D structured environments. We are also considering integrating tracking failure detection and re-localization mechanisms to deal with the kidnapping problem.

\balance

%%%%%%%%%%%%%%%%%%%%%%%%%%%%%%%%%%%%%%%%%%%%%%%%%%%%%%%
\bibliographystyle{IEEEtran}
\bibliography{icra2024_02}

\end{document}